\documentclass[letterpaper]{article} 
\usepackage{aaai2026}  
\usepackage{times}
\usepackage{helvet}
\usepackage{courier}
\usepackage[hyphens]{url}
\usepackage{graphicx}
\usepackage{booktabs}
\usepackage{multirow}
\usepackage[table]{xcolor}
\usepackage{arydshln}
\usepackage{pifont}

\usepackage{adjustbox}
\usepackage{amsmath}
\usepackage{amssymb}
\usepackage{bm}
\definecolor{c1}{HTML}{003371}
\usepackage{natbib}
\usepackage{caption}
\usepackage{cleveref}
\usepackage[most]{tcolorbox}
\definecolor{blue1}{HTML}{013371}
\newtcbtheorem[]{trainprompt}{Case}%
{colback=gray!00,colframe=blue1,fonttitle=\bfseries, left=.02in, right=.02in,bottom=.02in, top=.02in}{trainprompt}
\usepackage{algorithm}
\usepackage{algorithmic}
\usepackage{newfloat}
\usepackage{listings}
\DeclareCaptionStyle{ruled}{labelfont=normalfont,labelsep=colon,strut=off}
\floatstyle{ruled}
\newfloat{listing}{tb}{lst}{}
\floatname{listing}{Listing}
\title{VerifyBench: A Systematic Benchmark for Evaluating Reasoning Verifiers Across Domains}
\author{
    Xuzhao Li\textsuperscript{\rm 1,}\textsuperscript{\rm 2}\equalcontrib\thanks{Work done during Xuzhao’s internship.},
    Xuchen Li\textsuperscript{\rm 3}\equalcontrib,
    Shiyu Hu\textsuperscript{\rm 4},
    Yongzhen Guo\textsuperscript{\rm 2}\thanks{Corresponding Authors.},
    Wentao Zhang\textsuperscript{\rm 1}\footnotemark[3]
}
\affiliations{
    \textsuperscript{\rm 1}PKU,
    \textsuperscript{\rm 2}Ant Group,
    \textsuperscript{\rm 3}ZGCA,
    \textsuperscript{\rm 4}NTU\\
    xuzhaoli2001@gmail.com, xuchenli1030@gmail.com, yongzhen.gyz@antgroup.com, wentao.zhang@pku.edu.cn
}

\usepackage{bibentry}

\begin{document}

\maketitle

\begin{abstract}
Large language models (LLMs) increasingly rely on reinforcement learning (RL) to enhance their reasoning capabilities through feedback. A critical challenge is verifying the consistency of model-generated responses and reference answers, since these responses are often lengthy, diverse, and nuanced. Rule-based verifiers struggle with complexity, prompting the use of model-based verifiers. However, specialized verifiers lack flexibility, while general LLM judges can be inconsistent. Existing research primarily focuses on building better verifiers, yet a systematic evaluation of different types of verifiers' performance across domains remains lacking, severely constraining the reliable development of Reinforcement Learning with Verifiable Reward (RLVR). To address this, we propose VerifyBench--a cross-domain comprehensive benchmark for systematically evaluating verifiers. We construct about 4,000 expert-level questions covering mathematics, physics, chemistry, and biology. Each question is equipped with reference answers and diverse responses. The reliability of the evaluation is ensured through a rigorous annotation process conducted by a multidisciplinary expert team. We design a four-dimensional experimental framework to comprehensively compare the performance boundaries of specialized verifiers and general LLMs under combined conditions of extracted answers vs. complete responses, and short vs. long outputs. Our evaluation uncovers fundamental trade-offs in verifiers: while specialized verifiers achieve leading accuracy (the best model reaching 96.48\% in chemistry), they exhibit deficiencies in recall; general models show stronger inclusivity but unstable precision. More importantly, we discover verifiers' high sensitivity to input structure and inherent limitations in cross-domain generalization, providing critical insights into the bottlenecks of current verifier technology.
\end{abstract}

\section{Introduction}

\begin{figure}[t!]
  \centering   
  \includegraphics[width=\linewidth]{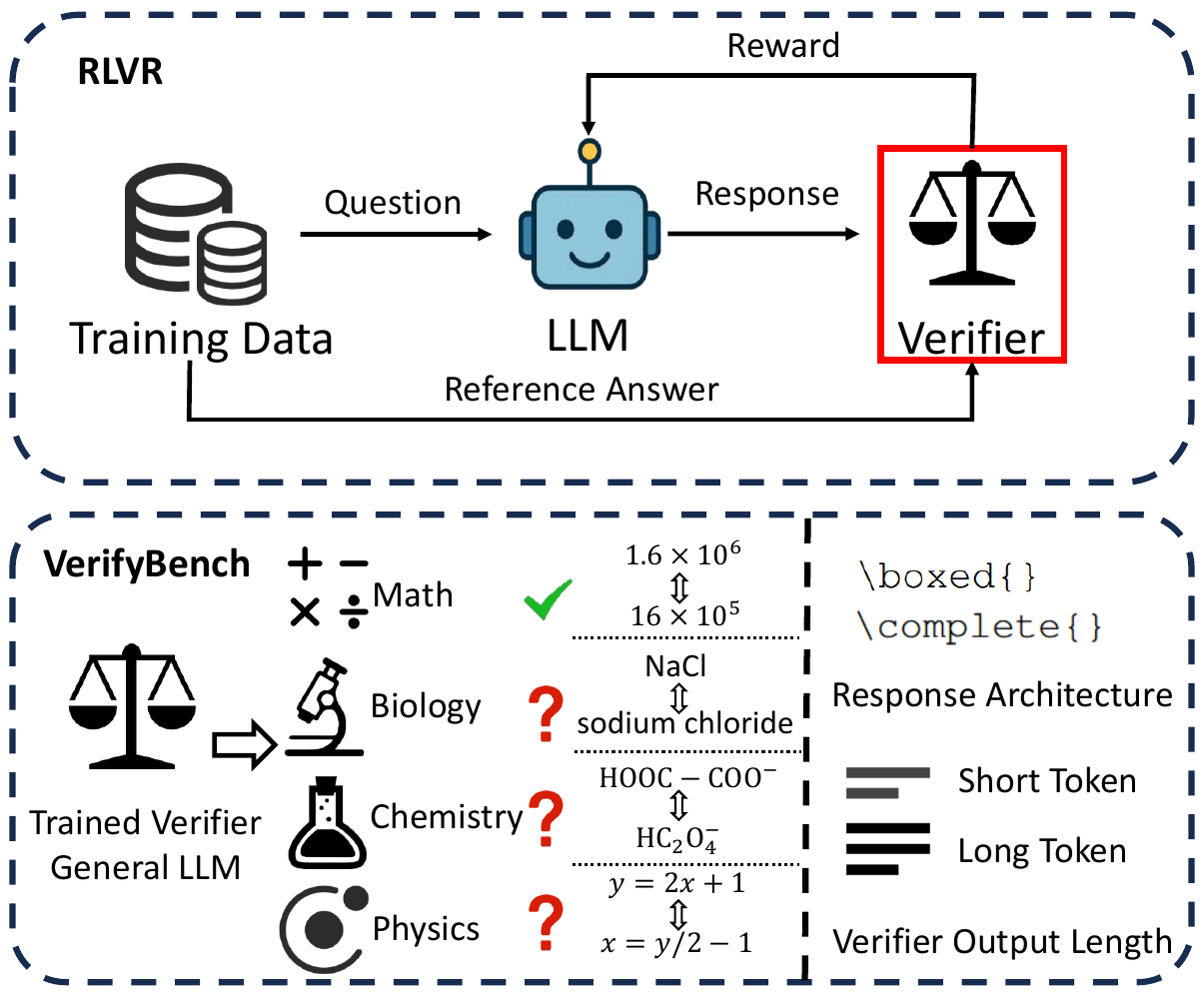}
  \caption{Overview of the Reinforcement Learning with Verifier (RLVR) paradigm and the VerifyBench evaluation framework. The upper section illustrates the verifier's role in the RL feedback loop for LLMs. The lower section highlights VerifyBench's multidisciplinary scope and the key experimental variables: different verifier types, response input formats (\texttt{\textbackslash boxed\{\}} vs. \texttt{\textbackslash complete\{\}}), and verifier output token lengths.}
  \label{fig:insight}
  \vspace{-10pt}
\end{figure}

Large language models (LLMs) \cite{deepseekr1,gpt4o,kimik1_5} have achieved significant breakthroughs in complex reasoning, planning, and symbolic problem-solving \cite{cao2025large}. This progress is largely driven by Reinforcement Learning with Verifiable Reward (RLVR) \cite{deepscaler2025,zeng2025simplerl}, which employs an external feedback loop to refine policy model. As depicted in Figure \ref{fig:insight}, this mechanism enables LLMs to generalize and produce high-quality responses. In recent work, this paradigm has been extended beyond rule-based reward \cite{math_verify_24_github} to include verifier-based learning \cite{xverify,ma2025general}, where model outputs are evaluated by a separate verifier model.

However, the inherent unreliability of current verifier systems remains a critical challenge \cite{huang2025pitfalls}. Rule-based verifiers, relying on rigid pattern matching, are limited in generalization and brittle with diverse, nuanced LLM responses. They frequently misclassify semantically correct but non-canonical responses, especially in complex domains like mathematics and science. This fundamental limitation creates misalignment between model performance and reward signals, directly hindering RLVR's scalability and trustworthiness in real-world deployment.

To overcome these constraints, model-based verifiers have emerged, including specialized models (finetuned on labeled data) and general-purpose LLMs \cite{qwen3,cai2024internlm2technicalreport,qwen2025qwen25technicalreport} acting as judges. While promising enhanced flexibility, their practical application introduces new issues. Specialized verifiers, though accurate, often lack adaptability to novel expressions. General-purpose LLMs frequently suffer from inconsistency and insufficient precision, particularly for step-by-step reasoning \cite{huang2025pitfalls}. Crucially, despite intense focus on verifier development, a systematic and comprehensive evaluation of diverse verifier types across varied domains and conditions remains absent. This significant gap impedes robust RLVR development, leaving practitioners without clear guidance.

To fill this critical void, we introduce VerifyBench: a cross-domain comprehensive benchmark for systematic evaluation of verifiers. VerifyBench is constructed from about 4,000 expert-level questions spanning mathematics, physics, chemistry, and biology. Each question includes reliable reference answers and diverse Chain-of-Thought (CoT) \cite{ma2025cot,wei2022chain} responses. Gold-standard judgment labels are established through a rigorous, fine-grained human annotation process by a multidisciplinary expert team, ensuring unparalleled reliability. This enables systematic analysis of verifier behavior across a four-dimensional experimental framework, varying input granularity (boxed final answers vs. full reasoning traces) and output constraints.

Our benchmark and experimental design are driven by three primary goals. First, to provide a systematic, multidisciplinary evaluation platform for verifiers across STEM domains. Second, to comprehensively analyze intricate behavioral differences between verifier types. Third, by simulating realistic RLVR deployment scenarios through varied input and output conditions, we uncover how these factors affect verifier reliability and expose critical bottlenecks in their current design.

Through controlled experiments, we rigorously evaluate both specialized verifiers (finetuned LLMs) and general-purpose LLMs (zero-shot or few-shot verifiers). Our findings reveal fundamental trade-offs in verifier design: specialized models offer leading precision but exhibit deficiencies in recall and struggle with diverse expressions. Conversely, general LLMs with larger model sizes demonstrate high inclusiveness but suffer from inconsistent structured judgment and a heightened risk of false positives. Importantly, we empirically uncover verifiers' acute sensitivity to input structure and inherent limitations in cross-domain generalization, providing critical insights into the fundamental challenges facing current verifiers. VerifyBench offers a rigorous foundation for developing trustworthy evaluations in RL-trained LLMs.

Our main contributions are as follows:
\begin{itemize}
\item We introduce VerifyBench, a novel, multidisciplinary benchmark of about 4,000 expert-level questions across mathematics, physics, chemistry, and biology, featuring fine-grained human annotations and diverse CoT responses from state-of-the-art LLM, specifically designed for systematic verifier evaluation.

\item We conduct a comprehensive empirical study comparing various verifier types---specialized models vs. general-purpose LLMs---under varying input contexts and output constraints. This systematic framework reveals fundamental judgmental trade-offs and highlights their limitations, particularly concerning precision, recall, and robustness to diverse reasoning.

\item We identify key challenges critical for verifier deployment, including the pervasive accuracy-recall trade-offs, input structure sensitivity, and cross-domain generalization limitations. Based on these insights, we propose actionable directions for designing more robust and generalizable verification systems to accelerate the reliable development of RLVR.
\end{itemize}

\section{Related Work}
\subsection{Reinforcement Learning with Verifiable Reward}
Reinforcement learning (RL) \cite{hu2025open,yu2025dapo,skywork-or1-2025,yue2025does} has emerged as a powerful paradigm for aligning LLMs with task-specific goals by optimizing reward signals through interaction. In tasks with well-defined outputs—such as mathematics \cite{hendrycks2021measuring}, programming \cite{jain2024livecodebench}, and logic puzzles \cite{xie2025logic}—automatic verifiers are frequently employed to provide reward feedback by evaluating the correctness of model responses. This technique enables scalable supervision without exhaustive human labeling and has been integrated into the training pipelines of many recent high-performing models \cite{li2025torl}. While effective in principle, the choice and reliability of the verifier are critical. Most existing systems rely on rule-based verifiers that operate by matching the model’s final answer against a reference \cite{li2025limr}, often using hand-crafted rules or symbolic equivalence criteria. These approaches are efficient but brittle: they may fail to recognize semantically correct answers expressed in alternative forms or with minor formatting differences. Furthermore, in multi-step reasoning scenarios, a strict match on the final boxed answer can overlook the model’s overall process quality. This creates a mismatch between verifier judgment and human evaluation standards, especially when the reasoning path is correct but the expression is unconventional.

\begin{figure*}[t!]
  \centering   
  \includegraphics[width=0.95\linewidth]{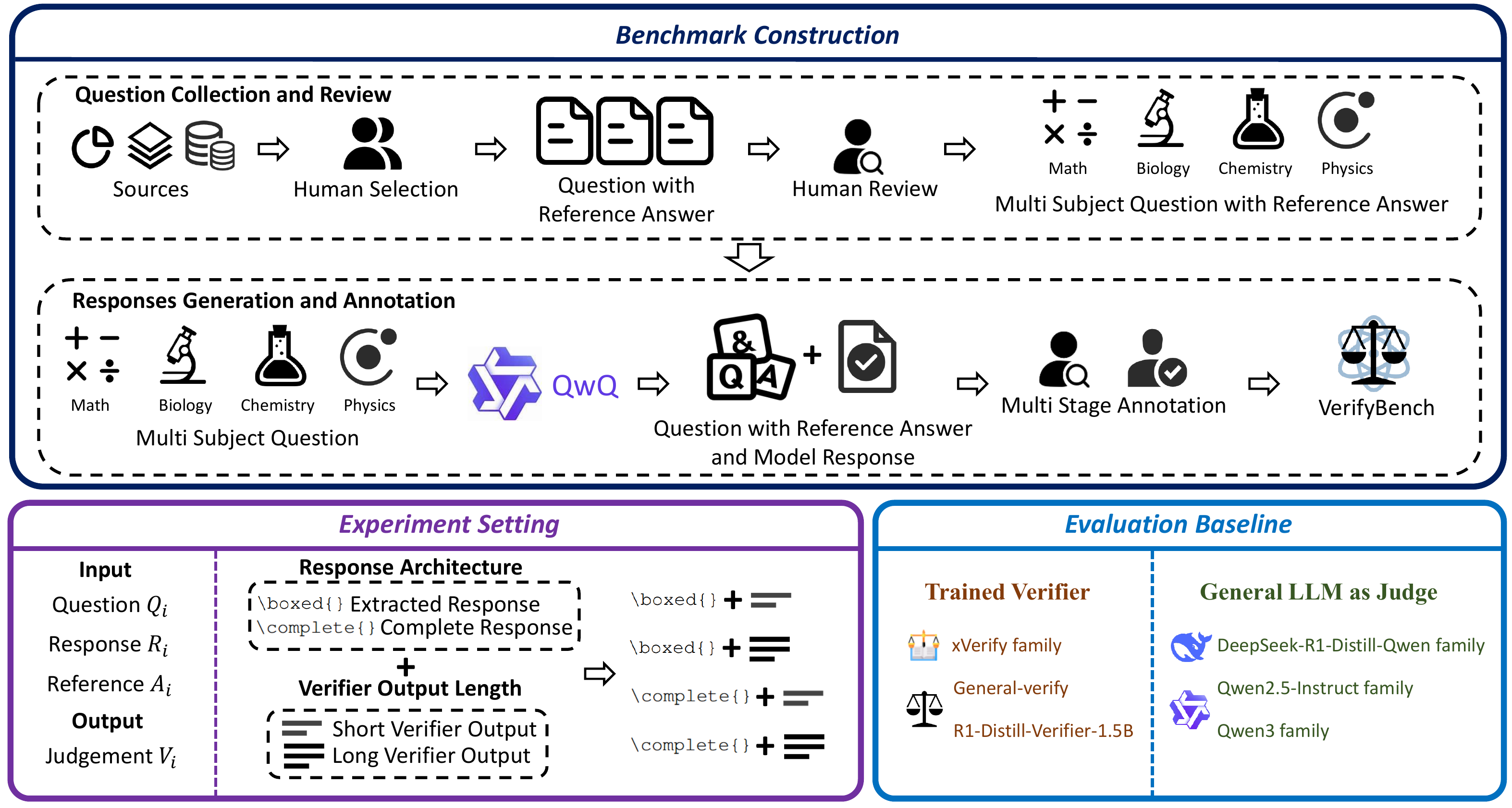}
  \caption{Overall framework of the VerifyBench. The diagram illustrates the meticulous benchmark construction process, including question collection, human review, response generation by QwQ-32B, and multi-stage human annotation. It also details the experimental settings, such as different response architectures (\texttt{\textbackslash boxed\{\}} vs. \texttt{\textbackslash complete\{\}}) and verifier output lengths (short vs. long), alongside the diverse categories of verifiers evaluated.}
  \label{fig:overall_framework}
  \vspace{-10pt}
\end{figure*}

\subsection{Model-based Verifier}
To address the limitations of static rule-based evaluation \cite{math_verify_24_github}, recent work explores trained verifier \cite{xverify,huang2025pitfalls} or general LLMs \cite{qwen2025qwen25technicalreport,yang2024qwen2math} as verifiers. Specialized verifiers are trained to predict whether a model-generated answer is valid based on responses and reference answers. These verifiers can handle richer linguistic and symbolic variation, and offer more flexible evaluation in tasks involving complex reasoning \cite{jiang2025deepretrieval}. However, trained verifiers may be overly tolerant, leading to reduced selectivity and accuracy \cite{ma2025general}. Some studies also propose using general LLMs as verifiers \cite{jin2025search}, leveraging their generalization capabilities and world knowledge in a few-shot or zero-shot setting. Despite their increasing use, a systematic understanding of how different types of verifiers perform under varying input and output conditions remains lacking. Our work addresses this gap through a unified benchmark and controlled comparison across verifier paradigms.

Crucially, while prior research has focused on developing specific verifier models, a comprehensive and systematic evaluation of their performance across diverse scenarios and domains is absent. Our work, VerifyBench, uniquely addresses this gap by establishing a rigorous benchmark and framework to assess existing and future verifiers, rather than proposing new architectures.

\section{Task Overview}

To verify the consistency between the model’s response and the reference answer, we formulate this as a binary classification problem \cite{huang2025pitfalls}. The verifier takes the question, the model’s response, and the reference answer as input. The model’s response is either the answer extracted from the model’s complex response (\texttt{\textbackslash boxed\{\}}) using a designated extraction function or the complete response (\texttt{\textbackslash complete\{\}}). Formally, each instance of the task is defined as a 4-tuple $(Q, R, A, V)$, where:

\begin{itemize}
    \item $Q_i$ denotes the $i$-th question.
    \item $R_i$ denotes the model's complete response to question.
    \item $A_i$ denotes the reference answer (i.e., ground truth).
\end{itemize}

For each instance $(Q_i, R_i, A_i)$, the verification process is formally defined by a verifier function $F_{\text{verifier}}$, which produces a binary evaluation result $V_i$:
\[
V_i = F_{\text{verifier}}(Q_i, R_i, A_i) = \mathcal{V}\big(\mathcal{H}(Q_i, R_i), A_i\big)
\]
where: $\mathcal{H}(Q_i, R_i)$ specifies the actual input provided to the core judgment function $\mathcal{V}$. It could be $\mathcal{H}(Q_i, R_i) = g(Q_i, R_i)$, where $g$ is an answer extraction function applied to the model response. For example, $g$ might extract the content inside the last \texttt{\textbackslash boxed\{\}} environment in $R_i$ as the model's final answer. In this case, the verifier primarily compares the extracted answer against the reference answer with the question. Alternatively, it could be $\mathcal{H}(Q_i, R_i) = (Q_i, R_i)$, meaning the verifier receives the original question and the full model response as context for its judgment. This allows the verifier to assess the reasoning process, not just the final answer.

$\mathcal{V}(\cdot, \cdot)$ is the binary judgment function. It takes the processed model output from $\mathcal{H}$ and the reference answer $A_i$ as input, returning $1$ if the model's response is deemed consistent with the reference answer, and $0$ otherwise.

\section{VerifyBench}

\subsection{Question Collection and Solution Generation}
We have carefully selected a total of approximately 4,000 high-quality, expert-level questions, covering four disciplines: mathematics, physics, chemistry, and biology, with around 1,000 questions per subject. These questions are designed to be challenging, incorporating rich domain-specific background, requiring advanced reasoning capabilities from LLMs, and allowing for open-ended, expressive answer spaces—thereby increasing benchmark complexity. Our definition of “expert-level” questions refers to problems typically encountered in university-level coursework or national/international academic competitions, characterized by their requirement for multi-step reasoning and synthesis of multiple concepts.

As shown in Figure \ref{fig:overall_framework}, we establish a dedicated data team composed of multidisciplinary experts with strong academic backgrounds in each respective domain to ensure data quality. The collection process strictly adheres to rigorous quality control standards and follows several key steps:

\begin{itemize}
    \item Drawing from both domestic and international sources, team members manually select problems that align with our “expert-level” definition. These questions are chosen to ensure broad topical coverage, high cognitive demand, and diversity in conceptual, theoretical, and applied reasoning, always paired with correct and clear reference answers. A detailed description of our question selection criteria and source identification process is provided in the Appendix.
    
    \item Each question undergoes 2 rounds of manual review. Cross-validation is conducted by graduate-level (Master's and PhD) reviewers to identify redundant knowledge points, overly simplistic phrasing, ambiguous wording, or irrelevant content. This careful review process ensures that the questions and reference answers are professional, coherent, and free from noise. Further specifics on our review protocols and reviewer qualifications can be found in the Appendix.
\end{itemize}

We use QwQ-32B~\cite{qwq32b} to generate detailed CoT responses for collected questions. The model is prompted to enclose its final answer in the \texttt{\textbackslash boxed\{\}}. These generated responses frequently contain self-reflective reasoning patterns and intermediate states, which present significant challenges to the verifier.

\subsection{Benchmark Annotation}

To ensure high-quality and consistent labeling of alignment between model responses and reference answers, we adopt a two-stage annotation process.

In the first stage, each instance (question, complete model response, and reference answer) is annotated by two annotators and reviewed by an independent reviewer. In the second stage, 200 questions are randomly sampled from each discipline (800 in total across 4 domains) for cross-validation by two annotators. Throughout the annotation process, we implement a real-time feedback mechanism: for complex, ambiguous, or disputed cases, annotators consult the data construction and standards team, reach a consensus through discussion, and document representative cases to iteratively refine the annotation guidelines.

For short-answer evaluation, annotation criteria include:

\begin{enumerate}
    \item Recognition of superficial answer variations, such as case insensitivity (``a'' vs. ``A'') or equivalence of ``alpha'' and ``$\alpha$''.
    \item Equivalence judgments across LaTeX expressions, symbolic formats, and natural language descriptions.
    \item Alignment assessment between the LLM's response (especially content inside \texttt{\textbackslash boxed\{\}}) and the ground truth, incorporating context and intermediate reasoning steps.
    \item Integrative judgment combining the reference answer’s explanatory context, question requirements, and the model’s response.
    \item Cases where the model derives the correct result but outputs an incorrect final answer are marked as wrong.
\end{enumerate}

Given the domain-specific nuances, we further extend and refine annotation principles as follows:

\paragraph{Mathematics and Physics:}
Attention is paid to units, notation, and equivalent solution paths. In physics, dimensional consistency and valid conversions are emphasized.

\paragraph{Chemistry:}
Chemical names (e.g., NaCl and sodium chloride), formulas, and states are treated as equivalent. Both International Union of Pure and Applied Chemistry (IUPAC) \cite{iupac1992international} and common names are accepted, with flexibility in reaction equation formats.

\paragraph{Biology:}
Variations in terminology (technical, vernacular, Latin) are tolerated if biologically accurate. Emphasis is placed on conceptual validity and mechanistic correctness.

\paragraph{Technical Formatting and Language:}
Differences in spelling, punctuation, and formatting are acceptable. In long-chain reasoning, correctness of logic and final answer are prioritized over surface-level language differences.

\subsection{Benchmark Statistics}

This section offers a detailed summary of VerifyBench, highlighting its scale, composition, and core characteristics across disciplines. The benchmark contains about 4,000 questions, each paired with a reference answer and a model-generated response, intended to support comprehensive and rigorous verifier evaluation. To ensure balanced representation, we include about 1,000 questions from each of four major scientific domains: mathematics, physics, chemistry, and biology. Key statistics are summarized in Table \ref{tab:benchmark_stats}.

\begin{table}[htbp!]
\centering
\caption{Key Statistics of VerifyBench}
\label{tab:benchmark_stats}
\begin{tabular}{ll}
\toprule
\textbf{Statistic} & \textbf{Value} \\
\midrule
Total Questions & 3,989 \\
Average Question Length & 186 tokens \\
Average Model Response Length & 4,553 tokens \\
\midrule
Total Annotated Instances & 3,989 \\
Label Distribution (Correct / Incorrect) & 45\% / 55\% \\
Inter-Annotator Agreement (IAA) & 0.88 -- 0.92 \\
\bottomrule
\end{tabular}
\end{table}

\begin{table*}[htbp!]
    \centering
    \begin{adjustbox}{max width=\textwidth}
    \resizebox{7in}{!}{
        \begin{tabular}{lcccccccccc}
          \toprule
          \multicolumn{1}{c}{\multirow{1}{*}{Verifier}} & Mathematics & Chemistry & Biology & Physics &  Overall \\
          \midrule
          \multicolumn{11}{c}{\textbf{Trained Verifier}} \\
          \midrule
          xVerify-0.5B-I  & 79.60\%\textbackslash 64.92\% & 94.77\%\textbackslash 96.50\% & 86.75\%\textbackslash \textbf{92.86\%} & 92.48\%\textbackslash 94.86\% & 88.77\%\textbackslash 88.66\% \\
          xVerify-3B-Ia  & 81.28\%\textbackslash 56.91\% & 94.77\%\textbackslash 96.21\% & 89.56\%\textbackslash 90.18\% & 91.57\%\textbackslash 92.24\% & 89.24\%\textbackslash 85.35\%  \\
          xVerify-8B-I  & 81.58\%\textbackslash 55.80\% & \textbf{96.39\%}\textbackslash \textbf{96.50\%} & 87.95\%\textbackslash 91.07\% & 92.68\%\textbackslash 92.69\% & 90.03\%\textbackslash 85.47\% \\ 
          xVerify-9B-C  & \textbf{83.03\%}\textbackslash 54.14\% & 96.28\%\textbackslash 94.46\% & \textbf{89.96\%}\textbackslash 91.07\% & 93.78\%\textbackslash 92.81\% & \textbf{90.96\%}\textbackslash 84.76\%  \\ 
          general-verify  & - & - & - & - & - \\ 
          R1-Distill-Verifier-1.5B  & - & - & - & - & - \\
          \midrule
          \multicolumn{11}{c}{\textbf{General LLM as Judge}} \\
          \midrule
          Qwen2.5-7B-Instruct & 78.18\%\textbackslash \textbf{91.99\%} & 89.45\%\textbackslash 82.22\% & 68.67\%\textbackslash 83.04\% & 86.66\%\textbackslash \textbf{98.17\%} & 83.52\%\textbackslash \textbf{93.74\%} \\ 
          Qwen2.5-14B-Instruct  &78.40\%\textbackslash 90.61\% & 90.05\%\textbackslash 82.80\% & 54.62\%\textbackslash 83.04\% & 91.28\%\textbackslash 96.46\% & 84.10\%\textbackslash 91.55\% \\ 
          Qwen2.5-32B-Instruct  & 80.98\%\textbackslash 76.24\% & 85.63\%\textbackslash 61.81\% & 61.45\%\textbackslash 66.96\% & \textbf{95.39\%}\textbackslash 97.26\% & 85.34\%\textbackslash 83.58\% \\  
          Qwen3-8B/14B/32B & - & - & - & - & - \\
          DS-R1-Distill-Qwen-7B/14B/32B & - & - & - & - & - \\
          \bottomrule
        \end{tabular}}
    \end{adjustbox}
    \caption{Performance comparison on VerifyBench including mathematics, chemistry, biology and physics. The table is organized by the trained verifier and the general LLM as the judge with the \texttt{\textbackslash complete\{\}} response from QwQ-32B, and the maximum output token size is set to 8. “-” means that in such a setup, the output of the verifier does not contain valid judgments. “DS” denotes DeepSeek. The best results are highlighted in bold.}
    \label{tab:main_results_2}
    \vspace{-10pt}
\end{table*}

\begin{figure}[t!]
  \centering   
  \includegraphics[width=1\linewidth]{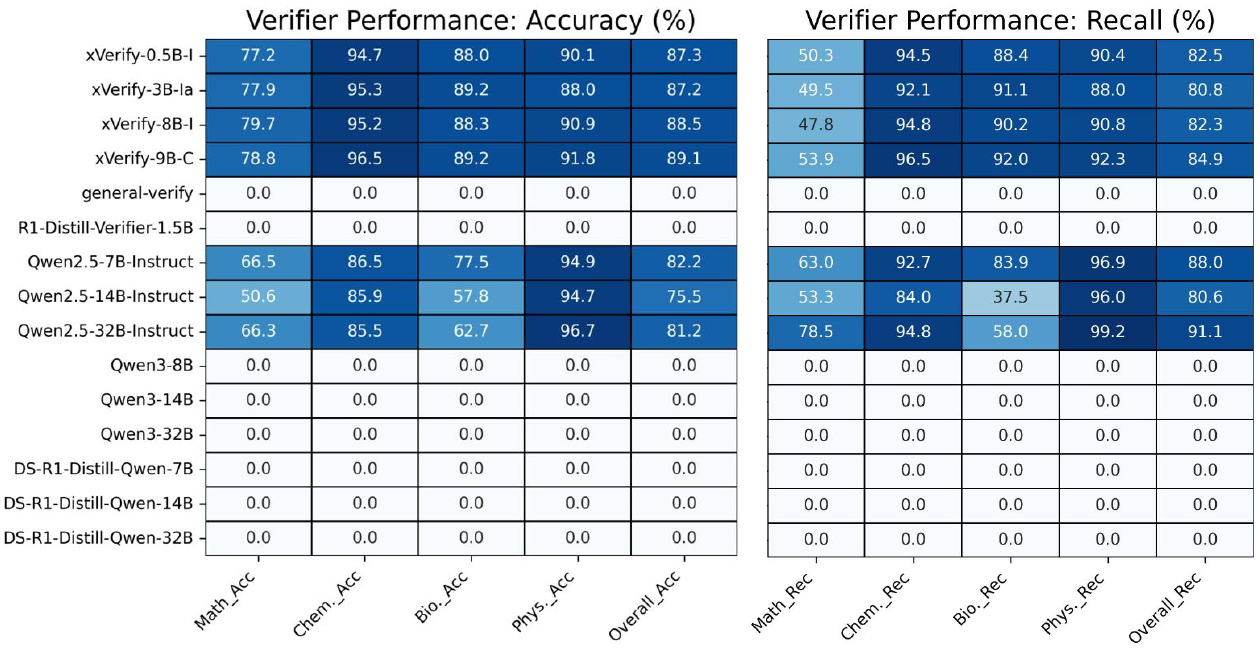}
    \caption{Performance comparison on VerifyBench including mathematics, chemistry, biology and physics. The figure is organized by the trained verifier and the general LLM as the judge with the response in the format \texttt{\textbackslash boxed\{\}} from QwQ-32B, and the maximum output token size is set to 8. 0.0 means that the output of the verifier does not contain valid judgments. “DS” denotes DeepSeek.}
    \label{tab:main_results_1}
  \vspace{-10pt}
\end{figure}

VerifyBench emphasizes long-form, multi-step reasoning, representing a substantial challenge for existing verifiers. The high average token count for responses (Table~\ref{tab:benchmark_stats}) reflects the benchmark’s focus on detailed, explanatory solutions and verbose model outputs. Our two-stage human annotation pipeline ensures high-quality binary labels, with inter-annotator agreement (IAA) \cite{artstein2017inter} scores ranging from 0.88 to 0.92, confirming annotation reliability.

Importantly, VerifyBench is designed to expose verifier limitations in both accuracy and robustness. It includes:
\begin{itemize}
    \item Semantically diverse correct responses, which require verifiers to go beyond string-level matching;
    \item Subtle reasoning errors and logical inconsistencies, testing fine-grained discrimination;
    \item Cross-domain coverage, requiring awareness of domain-specific notation, units, and reasoning conventions.
\end{itemize}

\begin{figure}[t!]
  \centering   
  \includegraphics[width=\linewidth]{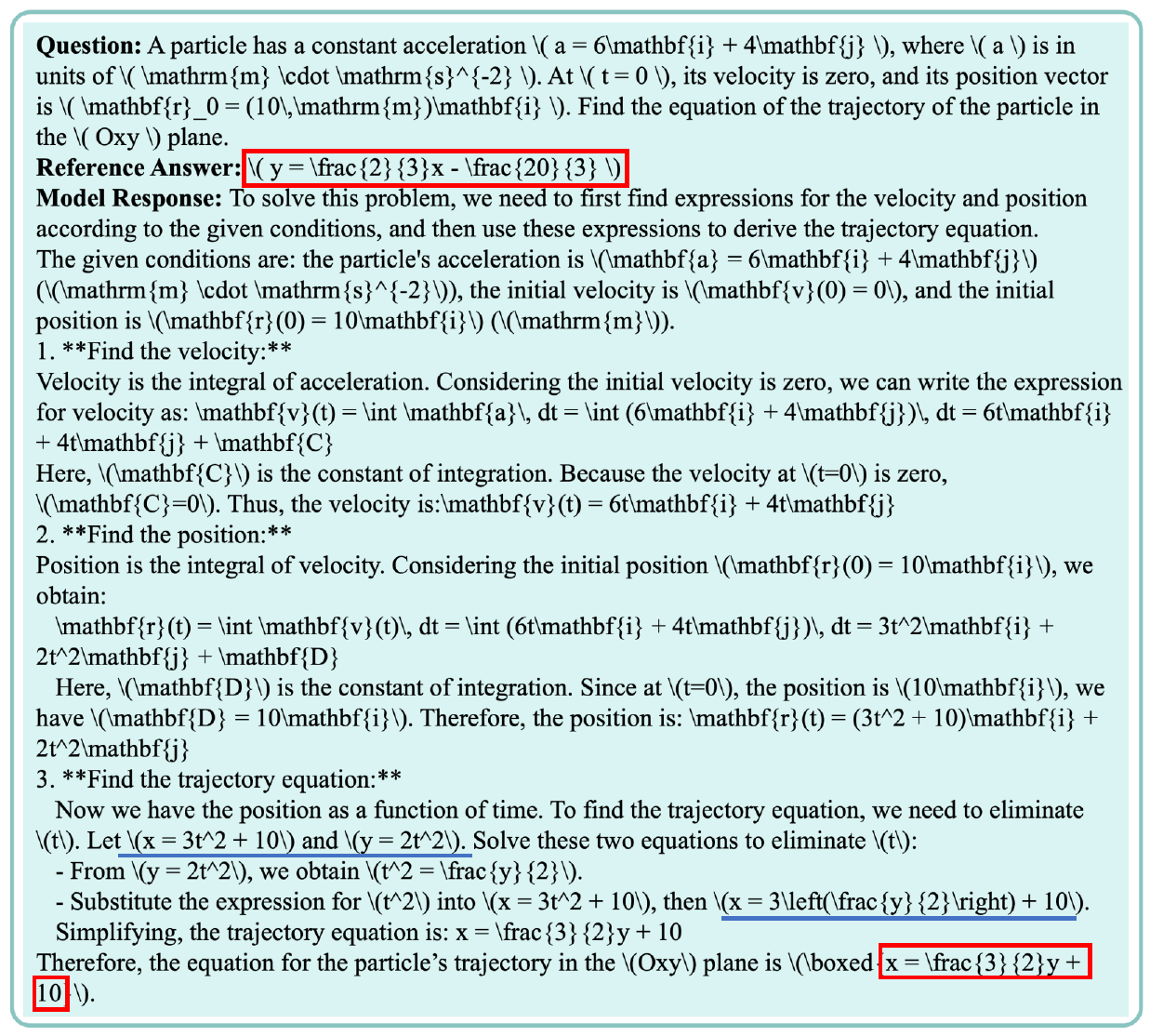}
  \vspace{-15pt}
  \caption{A physics case for evaluating the verifier's capabilities of precise answer extraction from long text, symbolic-equivalence simplification and comparison, tolerance to variable placement, and robust parsing of LaTeX with minor syntax errors.}
  \label{fig:phy2}
  \vspace{-10pt}
\end{figure}

Together, these characteristics create a rigorous and realistic evaluation setting that challenges verifiers to combine natural language understanding with domain-specific reasoning competence. We provide a physics case in Figure \ref{fig:phy2}. Appendix provides more detailed cases.

\begin{figure*}[t!]
  \centering   
  \vspace{-10pt}
  \includegraphics[width=0.95\linewidth]{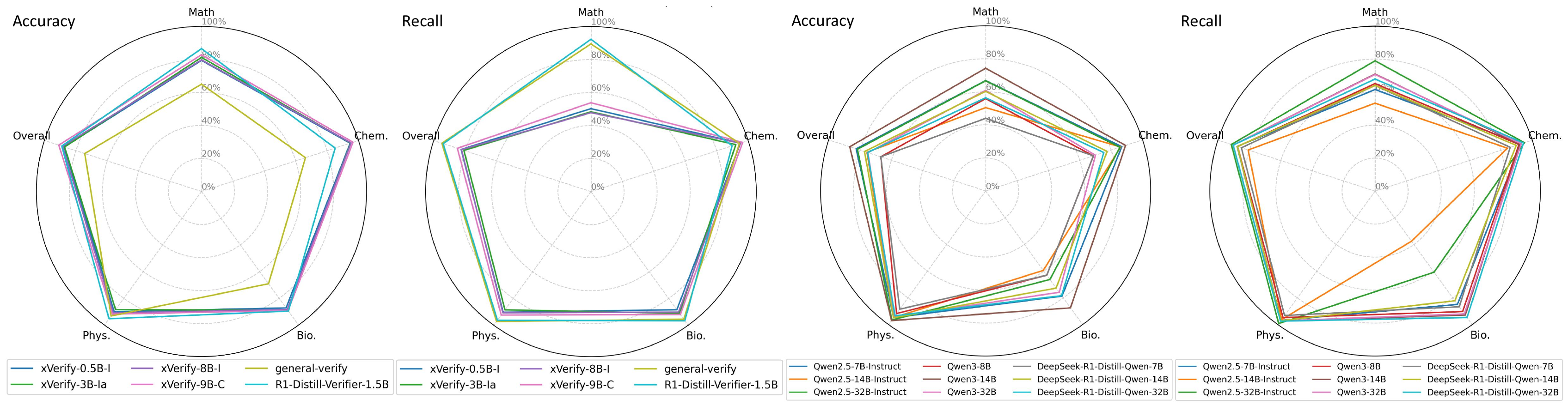}
    \caption{Performance comparison on VerifyBench including mathematics, chemistry, biology and physics. The figure is organized by the trained verifier (left) and the general LLM as the judge (right) with the response in the format \texttt{\textbackslash boxed\{\}} from QwQ-32B, and the maximum output token size is set to 4k. }
  \label{tab:main_results_3}
  \vspace{-10pt}
\end{figure*}

\section{Experiment}
\subsection{Settings}

\paragraph{Response Generation.} To simulate realistic response patterns of reasoning-capable language models, we employ QwQ-32B~\cite{qwq32b} to generate detailed CoT responses for all multi-subject questions in VerifyBench. We use the official recommended hyperparameters: temperature 0.6, TopP 0.95, MinP 0, TopK 40, and no repetition penalty. The maximum response length for QwQ-32B is set to 32,768 tokens.

\paragraph{Baseline.} We evaluate two categories of LLMs as verifiers. The first category comprises general-purpose open-source LLMs, including the Qwen2.5-Instruct series (7B, 14B, 32B) \cite{qwen2025qwen25technicalreport}, the Qwen3 series (8B, 14B, 32B) \cite{qwen3}, and the DeepSeek-R1-Distill-Qwen series (7B, 14B, 32B) \cite{deepseekr1}. The second category consists of specialized verifiers trained on large-scale verification datasets, such as the xVerify series (0.5B, 3B, 8B, 9B) \cite{xverify}, general\_verifier\cite{ma2025general}, and R1-Distill-Verifier-1.5B \cite{huang2025pitfalls}.

Among these, models from the xVerify and Qwen2.5-Instruct series directly output binary verification decisions (e.g., “correct” or “incorrect”). In contrast, generative models—such as the Qwen3 and DeepSeek-R1-Distill-Qwen series, as well as general\_verifier and R1-Distill-Verifier-1.5B—produce reasoning traces alongside final judgments. 

\paragraph{Verifier Inference.} The prompts for all models are unified as the prompt of xVerify \cite{xverify}. All models are evaluated under the same decoding configuration: temperature is set to 0.2, and TopP to 0.95. 

\paragraph{Evaluation Variants.} We consider the following four experimental settings, all of which include the question and reference answer as inputs:

\begin{itemize}
    \item \textbf{Boxed-only input, short output:} The verifier receives the extracted final answer (\texttt{\textbackslash boxed\{\}}), and the output is limited to 8 tokens.
    \item \textbf{Full-CoT input, short output:} The verifier receives the full model response (\texttt{\textbackslash complete\{\}}), with output again limited to 8 tokens.
    \item \textbf{Boxed-only input, long output:} The verifier receives only the extracted answer (\texttt{\textbackslash boxed\{\}}), but may generate up to 4k tokens.
    \item \textbf{Full-CoT input, long output:} The verifier receives the full response (\texttt{\textbackslash complete\{\}}) and may generate up to 4k tokens.
\end{itemize}

\subsection{Main Results}

\begin{table*}[t!]
    \centering
    \vspace{-10pt}
    \begin{adjustbox}{max width=\textwidth}
    \resizebox{6.8in}{!}{
        \begin{tabular}{lcccccccccc}
          \toprule
          \multicolumn{1}{c}{\multirow{1}{*}{Verifier}} & Mathematics & Chemistry & Biology & Physics &  Overall \\
          \midrule
          \multicolumn{11}{c}{\textbf{Trained Verifier}} \\
          \midrule
          xVerify-0.5B-I  & 79.38\%\textbackslash 54.14\% & 94.67\%\textbackslash 94.46\% & 87.15\%\textbackslash 91.07\% & 92.38\%\textbackslash 92.81\% & 88.64\%\textbackslash 84.76\% \\
          xVerify-3B-Ia  & 81.18\%\textbackslash 56.91\% & 95.03\%\textbackslash 96.21\% & 88.76\%\textbackslash 90.18\% & 91.88\%\textbackslash 92.24\% & 89.38\%\textbackslash 85.35\% \\ 
          xVerify-8B-I  & 81.58\%\textbackslash 55.80\% & \textbf{96.28\%}\textbackslash \textbf{96.50\%} & 89.16\%\textbackslash 91.07\% & 92.58\%\textbackslash 92.69\% & 90.06\%\textbackslash 85.47\% \\ 
          xVerify-9B-C  & 82.78\%\textbackslash 64.92\% & 96.18\%\textbackslash \textbf{96.50\%} & \textbf{89.96\%}\textbackslash \textbf{92.86\%} & 93.98\%\textbackslash 94.86\% & 90.86\%\textbackslash 88.66\% \\
          general-verify  & 68.77\%\textbackslash \textbf{88.12\%} & 75.13\%\textbackslash 88.63\% & 73.09\%\textbackslash 85.71\% & 94.32\%\textbackslash 97.72\% & 79.01\%\textbackslash 93.03\%\\ 
          R1-Distill-Verifier-1.5B  & 76.18\%\textbackslash 81.22\% & 80.71\%\textbackslash 86.30\% & 77.91\%\textbackslash 78.57\% & 88.77\%\textbackslash 89.50\% & 81.91\%\textbackslash 86.36\%\\
          \midrule
          \multicolumn{11}{c}{\textbf{General LLM as Judge}} \\
          \midrule
          Qwen2.5-7B-Instruct & 77.88\%\textbackslash 47.51\% & 89.45\%\textbackslash 93.59\% & 54.62\%\textbackslash 29.46\% & 86.66\%\textbackslash 86.64\% & 82.35\%\textbackslash 75.90\% \\ 
          Qwen2.5-14B-Instruct  & 78.08\%\textbackslash 59.12\% & 90.15\%\textbackslash 86.01\% & 62.25\%\textbackslash 34.82\% & 91.67\%\textbackslash 92.47\% & 84.75\%\textbackslash 80.21\% \\ 
          Qwen2.5-32B-Instruct  & 81.08\%\textbackslash 71.27\% & 85.28\%\textbackslash \textbf{96.50\%} & 62.25\%\textbackslash 48.21\% & 95.39\%\textbackslash 97.37\% & 81.20\%\textbackslash 88.36\% \\ 
          Qwen3-8B & 70.77\%\textbackslash 74.31\% & 88.04\%\textbackslash 93.88\% & 81.53\%\textbackslash 83.04\% & 95.09\%\textbackslash 97.37\% & 84.38\%\textbackslash 90.79\% \\ 
          Qwen3-14B  & \textbf{85.39\%}\textbackslash 80.11\% & 92.61\%\textbackslash 95.92\% & 84.34\%\textbackslash \textbf{92.86\%} & \textbf{96.99\%}\textbackslash 98.52\% & \textbf{91.11\%}\textbackslash \textbf{93.68\%} \\
          Qwen3-32B  & 74.67\%\textbackslash 70.72\% & 83.82\%\textbackslash 93.00\% & 83.40\%\textbackslash 91.96\% & 95.89\%\textbackslash 97.15\% & 84.69\%\textbackslash 90.31\% \\
          DS-R1-Distill-Qwen-7B & 64.66\%\textbackslash 74.59\% & 75.38\%\textbackslash 85.42\% & 74.80\%\textbackslash 77.68\% & 91.78\%\textbackslash 97.03\% & 77.07\%\textbackslash 88.72\% \\
          DS-R1-Distill-Qwen-14B  & 76.18\%\textbackslash 78.73\% & 81.06\%\textbackslash 85.13\% & 68.67\%\textbackslash 64.29\% & 95.69\%\textbackslash 97.26\% & 83.02\%\textbackslash 88.66\%\\ 
          DS-R1-Distill-Qwen-32B  & 72.37\%\textbackslash 78.73\% & 79.10\%\textbackslash \textbf{96.50\%} & 72.29\%\textbackslash 85.71\% & 96.59\%\textbackslash \textbf{99.43\%} & 81.85\%\textbackslash 93.50\% \\
          \bottomrule
        \end{tabular}}
    \end{adjustbox}
    \caption{Performance comparison on VerifyBench including mathematics, chemistry, biology and physics. The table is organized by the trained verifier and the general LLM as the judge with the \texttt{\textbackslash complete\{\}} response from QwQ-32B, and the maximum output token size is set to 4k. “DS” denotes DeepSeek. The best results are highlighted in bold.}
    \label{tab:main_results_4}
    \vspace{-10pt}
\end{table*}

\subsubsection{Boxed-only Input and Short Output.}
This setting focuses on verifying surface-level consistency using only the final answer (e.g., within \texttt{\textbackslash boxed\{\}}), while limiting the verifier's output length. Such a minimal setup tests the verifier’s basic decision-making and robustness to brevity. As shown in Figure \ref{tab:main_results_1}, specialized verifiers consistently outperform general-purpose LLMs across subjects, especially in domains with more variable or technical expressions (e.g., chemistry and biology). For example, xVerify-9B-C achieves leading accuracy in chemistry (96.48\%), biology (89.16\%), and physics (91.78\%), with an overall accuracy (89.07\%). In contrast, general LLMs often rely on literal string matching and exhibit lower generalization, leading to poorer performance in diverse subject areas. Notably, general LLMs with larger model sizes (e.g., Qwen2.5-32B-Instruct) show improvement in structured domains like physics, sometimes exceeding specialized models in recall.

\subsubsection{Full-CoT Input and Short Output.}
The verifier receives the full reasoning process as input but remains limited in output length. This setting challenges the model to efficiently extract conclusions from complex context. Specialized verifiers again show dominant performance, particularly in accuracy. For instance, xVerify-9B-C yields the best overall accuracy (90.96\%) and leads in most subjects. While general LLMs such as Qwen2.5-7B-Instruct demonstrate high recall (e.g., 91.99\% in math), they often trade precision for inclusiveness. This highlights a key contrast: general LLMs can tolerate diverse answer forms but struggle with precision. The details are shown in Table \ref{tab:main_results_2}.

\subsubsection{Boxed-only Input and Long Output.}
This setting introduces detailed verifier responses while maintaining a simplified input. It evaluates the reasoning and explanatory abilities under minimal input. Figure \ref{tab:main_results_3} displays the detailed results. Specialized verifiers like xVerify-9B-C and R1-Distill-Verifier-1.5B continue to lead in accuracy and show robust recall, with the latter achieving 93.98\% overall recall. While general models (e.g., Qwen3-14B) exhibit impressive recall in physics and chemistry, their accuracy often falls short due to misjudgment of semantically inconsistent responses.

\subsubsection{Full-CoT Input and Long Output.}
This most comprehensive setting approximates real-world judgment scenarios, allowing verifiers to access both rich context and full expressive capacity. Details are presented in Table \ref{tab:main_results_4}. Among all models, xVerify-9B-C again leads with top-tier accuracy across chemistry (96.18\%), biology (89.96\%), and physics (93.98\%). Its strong performance underscores the effectiveness of structured training for specialized verifiers. Meanwhile, Qwen3-14B, a general LLM, achieves comparable overall accuracy (91.11\%) and excels in recall for subjects like physics and biology. Nonetheless, specialized verifiers maintain an advantage in precision and consistency.

\subsection{Analysis}

\subsubsection{Performance Across Settings.}
We explore two input types (extracted vs. complete) and two output lengths (8 vs. 4k tokens), reflecting various levels of evaluation granularity. Across all configurations, specialized verifiers consistently achieve higher accuracy, especially in fields demanding strict semantic consistency. Richer input and longer output lead to performance gains, but recall improvements are modest—indicating a preference for precision over inclusiveness. For instance, xVerify models maintain high accuracy in various settings, demonstrating strong reliability. In contrast, general LLMs exhibit greater sensitivity to input/output conditions. They perform poorly when limited to short outputs and simple inputs but show substantial gains when given more context and freedom. Larger LLMs (e.g., Qwen3-14B/32B, DeepSeek-R1-Distill-Qwen-32B) exhibit huge recall improvements, sometimes surpassing specialized models. However, their accuracy remains less stable due to looser judgment standards and overgeneralization. 

\subsubsection{Strictness vs. Inclusiveness.}
Specialized verifiers prioritize correctness and reject ambiguous or loosely matched responses, aiming to reduce false positives. This yields high accuracy but may sacrifice recall, particularly when faced with valid answer variants. In contrast, general LLMs adopt a more inclusive stance, recognizing broader expression forms and redundant reasoning. While this boosts recall, it increases the risk of accepting incorrect answers.

\subsubsection{Towards End-to-End Judgment.}
From a system perspective, verifiers should ideally produce direct, structured outputs (e.g., “Correct”/“Incorrect”) without relying on the extraction of model response or verifier's judgment result. This reduces engineering overhead and minimizes error propagation. Specialized verifiers show promise in supporting this end-to-end evaluation paradigm. Their structured training enables them to focus on key conclusions within noisy or verbose outputs. Enhancing robustness to formatting and expression variance is a promising direction.

\subsubsection{Recommendations and Hybrid Strategies.}
Specialized verifiers can benefit from training augmentation with varied answer forms and noisy reasoning chains to boost generalization and recall. General LLMs should be guided toward more structured outputs and fine-tuned for domain-specific judgment accuracy. A hybrid pipeline is recommended: use general LLMs for high-recall coarse filtering, followed by specialized verifiers for precision filtering.

\section{Conclusion}
This benchmark presents a comprehensive evaluation of specialized verifiers and general LLMs under varying input-output constraints across multiple subjects. Results show that specialized verifiers consistently lead in accuracy, particularly in structured tasks, while general LLMs excel in recall and flexibility, especially when they have a larger model size and are given full input and output freedom. However, both approaches have trade-offs: specialized verifiers struggle with answer diversity, and general LLMs risk misjudgment due to less precise criteria and unstructured outputs. Moreover, answer format and extraction dependency impose challenges for practical deployment. To address these gaps, future models should aim for end-to-end judgment with minimal or no reliance on the extraction of model-generated answers and verifiers’ judgment results, while improving robustness to expression variability. We recommend enhancing generalization in specialized verifiers and guiding general LLMs toward structured outputs through fine-tuning.

\clearpage
\bibliography{aaai2026}

\begin{thebibliography}{30}
\providecommand{\natexlab}[1]{#1}

\bibitem[{Achiam et~al.(2023)Achiam, Adler, Agarwal, Ahmad, Akkaya, Aleman, Almeida, Altenschmidt, Altman, Anadkat et~al.}]{gpt4o}
Achiam, J.; Adler, S.; Agarwal, S.; Ahmad, L.; Akkaya, I.; Aleman, F.~L.; Almeida, D.; Altenschmidt, J.; Altman, S.; Anadkat, S.; et~al. 2023.
\newblock Gpt-4 technical report.
\newblock \emph{arXiv preprint arXiv:2303.08774}.

\bibitem[{Artstein(2017)}]{artstein2017inter}
Artstein, R. 2017.
\newblock Inter-annotator agreement.
\newblock \emph{Handbook of linguistic annotation}, 297--313.

\bibitem[{Cai et~al.(2024)Cai, Cao, Chen et~al.}]{cai2024internlm2technicalreport}
Cai, Z.; Cao, M.; Chen, H.; et~al. 2024.
\newblock InternLM2 Technical Report.
\newblock arXiv:2403.17297.

\bibitem[{Cao et~al.(2025)Cao, Men, Liu, Zhang, Li, Lin, Sui, Cao, Liu, and Zhao}]{cao2025large}
Cao, P.; Men, T.; Liu, W.; Zhang, J.; Li, X.; Lin, X.; Sui, D.; Cao, Y.; Liu, K.; and Zhao, J. 2025.
\newblock Large language models for planning: A comprehensive and systematic survey.
\newblock \emph{arXiv preprint arXiv:2505.19683}.

\bibitem[{Chen et~al.(2025)Chen, Yu, Wang, Zhang, Tang, Xiong, Li, Yang, and Li}]{xverify}
Chen, D.; Yu, Q.; Wang, P.; Zhang, W.; Tang, B.; Xiong, F.; Li, X.; Yang, M.; and Li, Z. 2025.
\newblock xverify: Efficient answer verifier for reasoning model evaluations.
\newblock \emph{arXiv preprint arXiv:2504.10481}.

\bibitem[{Guo et~al.(2025)Guo, Yang, Zhang, Song, Zhang, Xu, Zhu, Ma, Wang, Bi et~al.}]{deepseekr1}
Guo, D.; Yang, D.; Zhang, H.; Song, J.; Zhang, R.; Xu, R.; Zhu, Q.; Ma, S.; Wang, P.; Bi, X.; et~al. 2025.
\newblock Deepseek-r1: Incentivizing reasoning capability in llms via reinforcement learning.
\newblock \emph{arXiv preprint arXiv:2501.12948}.

\bibitem[{He et~al.(2025)He, Liu, Liu, Yan, Wang, Cheng, Zhang, Zhang, Xu, Shen, Li, Zeng, Wei, Cheng, An, Liu, and Zhou}]{skywork-or1-2025}
He, J.; Liu, J.; Liu, C.~Y.; Yan, R.; Wang, C.; Cheng, P.; Zhang, X.; Zhang, F.; Xu, J.; Shen, W.; Li, S.; Zeng, L.; Wei, T.; Cheng, C.; An, B.; Liu, Y.; and Zhou, Y. 2025.
\newblock Skywork Open Reasoner Series.
\newblock \url{https://capricious-hydrogen-41c.notion.site/Skywork-Open-Reaonser-Series\\-1d0bc9ae823a80459b46c149e4f51680}.
\newblock Notion Blog.

\bibitem[{Hendrycks et~al.(2021)Hendrycks, Burns, Kadavath, Arora, Basart, Tang, Song, and Steinhardt}]{hendrycks2021measuring}
Hendrycks, D.; Burns, C.; Kadavath, S.; Arora, A.; Basart, S.; Tang, E.; Song, D.; and Steinhardt, J. 2021.
\newblock Measuring mathematical problem solving with the math dataset.
\newblock \emph{arXiv preprint arXiv:2103.03874}.

\bibitem[{Hu et~al.(2025)Hu, Zhang, Han, Jiang, Zhang, and Shum}]{hu2025open}
Hu, J.; Zhang, Y.; Han, Q.; Jiang, D.; Zhang, X.; and Shum, H.-Y. 2025.
\newblock Open-Reasoner-Zero: An Open Source Approach to Scaling Up Reinforcement Learning on the Base Model.
\newblock \emph{arXiv preprint arXiv:2503.24290}.

\bibitem[{Huang et~al.(2025)Huang, Zeng, Zeng, Zhu, and He}]{huang2025pitfalls}
Huang, Y.; Zeng, W.; Zeng, X.; Zhu, Q.; and He, J. 2025.
\newblock Pitfalls of Rule-and Model-based Verifiers--A Case Study on Mathematical Reasoning.
\newblock \emph{arXiv preprint arXiv:2505.22203}.

\bibitem[{Hynek~Kydlíček(2024)}]{math_verify_24_github}
Hynek~Kydlíček, G.~G. 2024.
\newblock {GitHub - huggingface/Math-Verify: A robust mathematical expression evaluation system designed for assessing Large Language Model outputs in mathematical tasks.}

\bibitem[{IUPAC(1992)}]{iupac1992international}
IUPAC, O. 1992.
\newblock International union of pure and applied chemistry.
\newblock \emph{Standard methods for the analysis of oils, fats and derivates}.

\bibitem[{Jain et~al.(2024)Jain, Han, Gu, Li, Yan, Zhang, Wang, Solar-Lezama, Sen, and Stoica}]{jain2024livecodebench}
Jain, N.; Han, K.; Gu, A.; Li, W.-D.; Yan, F.; Zhang, T.; Wang, S.; Solar-Lezama, A.; Sen, K.; and Stoica, I. 2024.
\newblock Livecodebench: Holistic and contamination free evaluation of large language models for code.
\newblock \emph{arXiv preprint arXiv:2403.07974}.

\bibitem[{Jiang et~al.(2025)Jiang, Lin, Cao, Tian, Kang, Wang, Sun, and Han}]{jiang2025deepretrieval}
Jiang, P.; Lin, J.; Cao, L.; Tian, R.; Kang, S.; Wang, Z.; Sun, J.; and Han, J. 2025.
\newblock Deepretrieval: Hacking real search engines and retrievers with large language models via reinforcement learning.
\newblock \emph{arXiv preprint arXiv:2503.00223}.

\bibitem[{Jin et~al.(2025)Jin, Zeng, Yue, Yoon, Arik, Wang, Zamani, and Han}]{jin2025search}
Jin, B.; Zeng, H.; Yue, Z.; Yoon, J.; Arik, S.; Wang, D.; Zamani, H.; and Han, J. 2025.
\newblock Search-r1: Training llms to reason and leverage search engines with reinforcement learning.
\newblock \emph{arXiv preprint arXiv:2503.09516}.

\bibitem[{Li, Zou, and Liu(2025{\natexlab{a}})}]{li2025limr}
Li, X.; Zou, H.; and Liu, P. 2025{\natexlab{a}}.
\newblock Limr: Less is more for rl scaling.
\newblock \emph{arXiv preprint arXiv:2502.11886}.

\bibitem[{Li, Zou, and Liu(2025{\natexlab{b}})}]{li2025torl}
Li, X.; Zou, H.; and Liu, P. 2025{\natexlab{b}}.
\newblock Torl: Scaling tool-integrated rl.
\newblock \emph{arXiv preprint arXiv:2503.23383}.

\bibitem[{Luo et~al.(2025)Luo, Tan, Wong, Shi, Tang, Roongta, Cai, Luo, Li, Popa, and Stoica}]{deepscaler2025}
Luo, M.; Tan, S.; Wong, J.; Shi, X.; Tang, W.~Y.; Roongta, M.; Cai, C.; Luo, J.; Li, L.~E.; Popa, R.~A.; and Stoica, I. 2025.
\newblock DeepScaleR: Surpassing O1-Preview with a 1.5B Model by Scaling RL.
\newblock \url{https://pretty-radio-b75.notion.site/DeepScaleR-Surpassing-O1-Preview-with-a-1-5B-Model-by-Scaling-RL-19681902c1468005bed8ca303013a4e2}.
\newblock Notion Blog.

\bibitem[{Ma et~al.(2025{\natexlab{a}})Ma, Liu, Jiang, Zhang, Ma, and Chen}]{ma2025general}
Ma, X.; Liu, Q.; Jiang, D.; Zhang, G.; Ma, Z.; and Chen, W. 2025{\natexlab{a}}.
\newblock General-reasoner: Advancing llm reasoning across all domains.
\newblock \emph{arXiv preprint arXiv:2505.14652}.

\bibitem[{Ma et~al.(2025{\natexlab{b}})Ma, Wan, Yu, Fang, and Wang}]{ma2025cot}
Ma, X.; Wan, G.; Yu, R.; Fang, G.; and Wang, X. 2025{\natexlab{b}}.
\newblock CoT-Valve: Length-Compressible Chain-of-Thought Tuning.
\newblock \emph{arXiv preprint arXiv:2502.09601}.

\bibitem[{Qwen et~al.(2025)Qwen, :, Yang, Yang, and Others}]{qwen2025qwen25technicalreport}
Qwen; :; Yang, A.; Yang, B.; and Others. 2025.
\newblock Qwen2.5 Technical Report.
\newblock arXiv:2412.15115.

\bibitem[{Team et~al.(2025)Team, Du, Gao, Xing, Jiang, Chen, Li, Xiao, Du, Liao et~al.}]{kimik1_5}
Team, K.; Du, A.; Gao, B.; Xing, B.; Jiang, C.; Chen, C.; Li, C.; Xiao, C.; Du, C.; Liao, C.; et~al. 2025.
\newblock Kimi k1. 5: Scaling reinforcement learning with llms.
\newblock \emph{arXiv preprint arXiv:2501.12599}.

\bibitem[{Team(2025)}]{qwq32b}
Team, Q. 2025.
\newblock QwQ-32B: Embracing the Power of Reinforcement Learning.

\bibitem[{Wei et~al.(2022)Wei, Wang, Schuurmans, Bosma, Xia, Chi, Le, Zhou et~al.}]{wei2022chain}
Wei, J.; Wang, X.; Schuurmans, D.; Bosma, M.; Xia, F.; Chi, E.; Le, Q.~V.; Zhou, D.; et~al. 2022.
\newblock Chain-of-thought prompting elicits reasoning in large language models.
\newblock \emph{Advances in neural information processing systems}, 35: 24824--24837.

\bibitem[{Xie et~al.(2025)Xie, Gao, Ren, Luo, Hong, Dai, Zhou, Qiu, Wu, and Luo}]{xie2025logic}
Xie, T.; Gao, Z.; Ren, Q.; Luo, H.; Hong, Y.; Dai, B.; Zhou, J.; Qiu, K.; Wu, Z.; and Luo, C. 2025.
\newblock Logic-rl: Unleashing llm reasoning with rule-based reinforcement learning.
\newblock \emph{arXiv preprint arXiv:2502.14768}.

\bibitem[{Yang et~al.(2025)Yang, Li, Yang, Zhang, Hui, Zheng, Yu, Gao, Huang, Lv et~al.}]{qwen3}
Yang, A.; Li, A.; Yang, B.; Zhang, B.; Hui, B.; Zheng, B.; Yu, B.; Gao, C.; Huang, C.; Lv, C.; et~al. 2025.
\newblock Qwen3 technical report.
\newblock \emph{arXiv preprint arXiv:2505.09388}.

\bibitem[{Yang et~al.(2024)Yang, Zhang, Hui, Gao, Yu, Li, Liu, Tu, Zhou, Lin et~al.}]{yang2024qwen2math}
Yang, A.; Zhang, B.; Hui, B.; Gao, B.; Yu, B.; Li, C.; Liu, D.; Tu, J.; Zhou, J.; Lin, J.; et~al. 2024.
\newblock Qwen2. 5-math technical report: Toward mathematical expert model via self-improvement.
\newblock \emph{arXiv preprint arXiv:2409.12122}.

\bibitem[{Yu et~al.(2025)Yu, Zhang, Zhu, Yuan, Zuo, Yue, Fan, Liu, Liu, Liu et~al.}]{yu2025dapo}
Yu, Q.; Zhang, Z.; Zhu, R.; Yuan, Y.; Zuo, X.; Yue, Y.; Fan, T.; Liu, G.; Liu, L.; Liu, X.; et~al. 2025.
\newblock Dapo: An open-source llm reinforcement learning system at scale.
\newblock \emph{arXiv preprint arXiv:2503.14476}.

\bibitem[{Yue et~al.(2025)Yue, Chen, Lu, Zhao, Wang, Song, and Huang}]{yue2025does}
Yue, Y.; Chen, Z.; Lu, R.; Zhao, A.; Wang, Z.; Song, S.; and Huang, G. 2025.
\newblock Does reinforcement learning really incentivize reasoning capacity in llms beyond the base model?
\newblock \emph{arXiv preprint arXiv:2504.13837}.

\bibitem[{Zeng et~al.(2025)Zeng, Huang, Liu, Liu, He, Ma, and He}]{zeng2025simplerl}
Zeng, W.; Huang, Y.; Liu, Q.; Liu, W.; He, K.; Ma, Z.; and He, J. 2025.
\newblock Simplerl-zoo: Investigating and taming zero reinforcement learning for open base models in the wild.
\newblock \emph{arXiv preprint arXiv:2503.18892}.

\end{thebibliography}

\end{document}